\documentclass{article}

\usepackage{arxiv}

\usepackage[utf8]{inputenc} 
\usepackage[T1]{fontenc}    
\usepackage{hyperref}       
\usepackage{url}            
\usepackage{booktabs}       
\usepackage{amsfonts}       
\usepackage{nicefrac}       
\usepackage{microtype}      
\usepackage{lipsum}		
\usepackage{graphicx}
\usepackage{natbib}
\usepackage{doi}
\usepackage{multirow}
\usepackage{bbding}
\usepackage{amsmath}
\usepackage{makecell}
\usepackage{tabularx}
\usepackage{subcaption}
\usepackage{booktabs}
\usepackage[dvipsnames]{xcolor}
\usepackage{tcolorbox}

\usepackage{listings} 
\usepackage{tikz}
\usepackage{enumitem}
\usepackage{colortbl}
\usepackage{adjustbox}

\newcommand{\modelname}{\textbf{InternVideo2.5}}
\def\checkmark{\tikz\fill[scale=0.4](0,.35) -- (.25,0) -- (1,.7) -- (.25,.15) -- cycle;} 

\title{{\modelname}: Empowering Video MLLMs with Long and Rich Context Modeling}

\author{
Yi Wang$^{*1,3}$, Xinhao Li$^{*2,1}$, Ziang Yan$^{*1}$, Yinan He$^{*1}$, Jiashuo Yu$^{*1}$\\
\textbf{Xiangyu Zeng$^{2,1}$, Chenting Wang$^{1}$, Changlian Ma$^{2,1}$, Haian Huang$^{1}$}\\
\textbf{Jianfei Gao$^{1}$, Min Dou$^{1}$, Kai Chen$^{1}$, Wenhai Wang$^{1}$}\\
\textbf{Yu Qiao$^{\dagger 1}$, Yali Wang$^{\dagger 4,1}$, Limin Wang$^{\dagger 2,1}$}\\\
 $^1$Shanghai AI Laboratory \quad $^2$Nanjing University \quad $^3$Shanghai Innovation Institute\\
 $^4$Shenzhen Institutes of Advanced Technology, Chinese Academy of Sciences \\ \\
 \small{\url{https://github.com/OpenGVLab/InternVideo/tree/main/InternVideo2.5}} \\
 \hspace{-0.25cm}
}

\definecolor{darkGreen}{RGB}{92, 148, 110}
\definecolor{myblue}{RGB}{14, 121, 178}

\renewcommand{\shorttitle}{{\modelname}: Empowering Video MLLMs with Long and Rich Context Modeling}

\renewcommand{\cite}{\citep}

\hypersetup{
pdftitle={{\modelname}: Empowering Video MLLMs with Long and Rich Context Modeling},
pdfsubject={computer vision, machine learning},
pdfauthor={Shanghai AI Lab},
pdfkeywords={Video representation, Multimodal learning, },
}

\begin{document}
\maketitle

\begin{figure*}[th]
    \centering
    \includegraphics[width=1\textwidth]{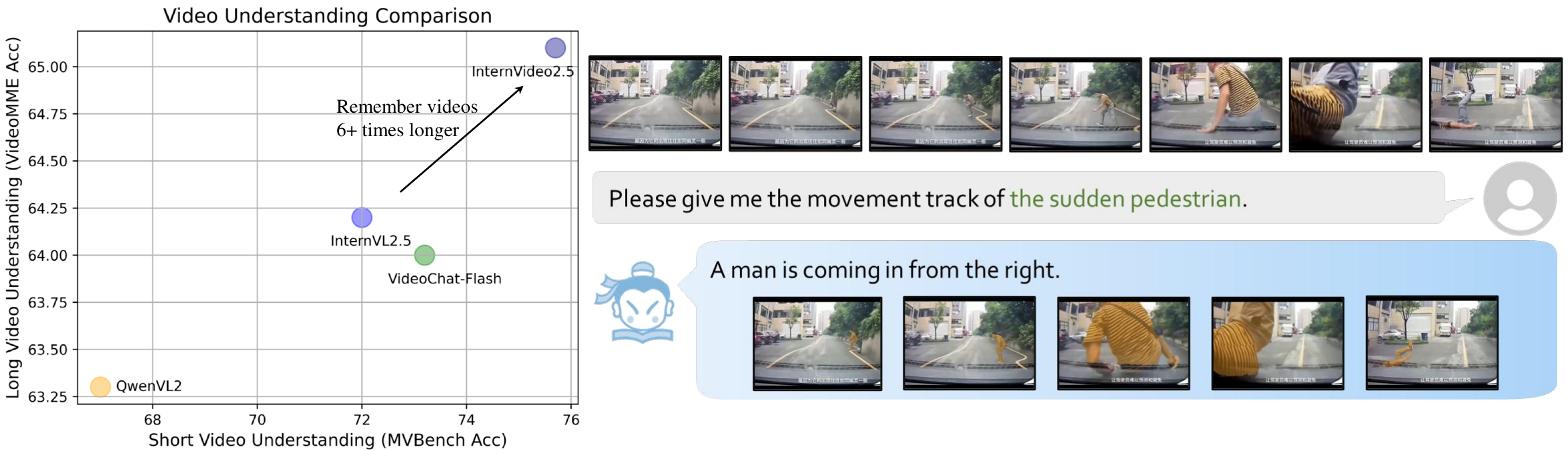}
    \vspace{-0.3cm}
    \caption{Demonstrations of {\modelname}. Left: open-source model (8B) performance on MVBench and VideoMME; right: an example of {\modelname} about it monitors the target requested by users and analyzes it.}
    \label{fig:teaser}
\end{figure*}
\begin{abstract}
{
This paper aims to improve the performance of video multimodal large language models (MLLM) via long and rich context (LRC) modeling. As a result, we develop a new version of {\modelname} with a focus on enhancing the original MLLMs' ability to perceive fine-grained details and capture long-form temporal structure in videos. Specifically, our approach incorporates dense vision task annotations into MLLMs using direct preference optimization and develops compact spatiotemporal representations through adaptive hierarchical token compression. Experimental results demonstrate this unique design of LRC greatly improves the results of video MLLM in mainstream video understanding benchmarks (short \& long), enabling the MLLM to memorize significantly longer video inputs (at least 6x longer than the original), and master specialized vision capabilities like object tracking and segmentation. Our work highlights the importance of multimodal context richness (length and fineness) in empowering MLLM's innate abilites (focus and memory), providing new insights for future research on video MLLM.
  \renewcommand{\thefootnote}%
    {\fnsymbol{footnote}}
  \footnotetext[0]{*Equal contribution. $\dagger$Corresponding authors.} 
  }
\end{abstract}

\section{Introduction}
Multimodal large language models (MLLM) make milestones in reaching artificial general intelligence. They integrate different sensing signals into a unified LLM-based framework, reformulating most perception and cognition problems into the multimodal next token prediction task. This successfully addresses applications ranging from multimodal document analysis \cite{gpt4o,gemini,internvl2}, video understanding \cite{videochat,internvideo2,li2024videochat}, agent interactions, to scientific discovery~\cite{chen2023fengwu}, world modeling \cite{agarwal2025cosmos}, autonomous driving~\cite{hu2023planning}, and live assistance~\cite{mu2024embodiedgpt, palme}. 

Despite its advances, MLLMs still underperform humans in fundamental vision related tasks, impeding its understanding and reasoning performance. They frequently exhibit difficulties in accurately recognizing, localizing, and recall objects, scenes, and motions in common scenarios - limitations that users find difficult to accept. Although research has demonstrated that scaling laws apply to vision-language modeling, showing consistent improvements in multimodal understanding benchmarks through increased visual-related data and model size, this trend doesn't provide a clear timeline for MLLMs to achieve human-level visual understanding. Current research emphasizes intelligent document analysis and resolution adaptation for high-definition processing. While these approaches show promising performance improvements, they don't conclusively demonstrate emerging capabilities in systematically solving these problems.

In this paper, we investigate how the length and fineness of the multimodal context influences MLLMs' vision-centric abilities and performance, rather than focusing on directly scaling MLLMs through model size or data volume. Here length and fineness refers to the model's ability to process and interpret multimodal inputs (e.g., video frames, audio, text) in long-term context with fine-grained details. Intuitively, this directly impacts the model's understanding and reasoning. Longer context allows the model to capture extended temporal dependencies and coherence, such as story arcs in videos or multi-step events. This is crucial for understanding complex narratives or reasoning over them. Meanwhile, fine-grained context, such as object details, spatiotemporal relationships, enables the model to perceive subtle details. It improves understanding specific actions, interactions, or scenes, and short-term reasoning on inferring causality or predicting future actions.

As higher context resolution and richness (both length and fineness) improves the model's ability to perceive and encode multimodal inputs accurately, we explore enhancing MLLMs by explicitly modeling fine-grained and extended multimodal context in a scalable manner.
For accurate spatiotemporal understanding, we transfer extensive dense visual annotations into MLLMs using direct preference optimization (DPO) \cite{rafailov2024dpo}, utilizing vision expert models as preference models \cite{yan2024task}. To process extended multimodal context, we employ compact and compatible spatiotemporal representations by adaptively compressing multimodal tokens both visually and semantically \cite{li2024videochat}.

Generally, we show that improving multimodal context length and fineness is a workable and easily implemented solution with quick returns in both chat-based and fundamental vision perception performance, as given in Figure. \ref{fig:teaser}. Considering whether the online interaction of MLLM more or less replies on its memory (how long its multimodal context can handle) and focus (how accurate its context can capture), we believe {\modelname} builds capability foundations for these advanced functions and applications. Concretely, our contributions lie in:
\begin{itemize}[leftmargin=*]
    \item To our best knowledge, we give a first comprehensive study on how to realize long and rich context (LRC) for improving MLLM's memory and focus. By unifying hierarchical token compression (HiCo) and task preference optimization (TPO) into a schema, we can enhance current MLLMs at scale with highly efficient learning and rich vision annotations.
    \item {\modelname} can improve existing MLLMs with notable performance in video understanding as well as granting them with expert-level visual perception abilities. Specifically, {\modelname} achieves the leading performance in several short and long video benchmarks. The video memory capacity of {\modelname} has been significantly enhanced, allowing it to retain inputs at least six times longer than the original.
\end{itemize}

\section{Related Work}
\paragraph{Multimodal Large Language Models.} Multimodal Large Language Models (MLLMs) 
usually combine vision encoder, LLM, and their connectors through instruction-tuning, capable of handling sophisticated tasks such as generating accurate picture descriptions and responding to visual queries. Progresses have been made to explore its model architecture \cite{zohar2024apollo,stllm,videochat,wang2022internvideo,mplugowl,ye2024mplug}, size \cite{internvl2,llavanextvideo} and capability \cite{wang2024visionllm}, training corpus \cite{li2024omnicorpus,wang2023internvid,glm2024chatglm}, preference optimization \cite{yan2024task,yu2024rlhf}, and more. Contemporary video-capable MLLMs have demonstrated the ability to process sequential visuals, comprehending spatiotemporal variations. Works discuss different how different vision encoders \cite{zohar2024apollo} and connectors \cite{stllm} affect MLLMs' performance, and find video encoders are still irreplaceable with image ones. Also, though MLPs related connectors are proven as effective as Q-Former, pooling, and other compression-preferred ones, the latter ones can deliver more efficient solutions, benefiting potential long visuals processing \cite{li2024videochat}. 

\paragraph{Long Video Understanding.} Long video understanding has seen rapid progress through multimodal large language models (MLLMs). It faces challenges in handling extended video content, leading researchers to pursue three primary approaches. The first method~\cite{gemini, uneven, longvila, longva} focuses on expanding the context window capacity of MLLM to handle longer sequences. The second emphasizes efficient token compression to reduce compute~\cite{llamavid,videoccam,longvlm,koala,moviechat,videoxl}. The third leverages agents to manage compute complexity, minaly disentangling long video understanding into temporal grounding, spatiotemporal perception, and the reasoning over the observed evidence~\cite{fan2025videoagent, wang2025videoagent}.

Context window extension has demonstrated promising results in processing longer video sequences. For instance, advanced training systems like parallel computing~\cite{2023xtuner} have been developed to optimize compute distribution across temporal and tensor dimensions, improving computational scalability by using more devices. However, while these approaches~\cite{gemini, uneven, longvila, longva} enable the processing of longer videos, they often encounter barriers due to communication costs and real video length.

Token compression creates compact video representations while attempting to preserve essentials. These methods~\cite{llamavid,videoccam,longvlm, koala,moviechat,videoxl} achieve high compression ratios, making them more computationally efficient. However, their performance in detailed video understanding tasks often falls short, sometimes performing below image-focused MLLMs, suggesting room for improvement in maintaining semantics during compression.

Agent-based methods decompose long video comprehension into several sub-tasks for existing expert models and LLMs. These methods~\cite{fan2025videoagent, wang2025videoagent, fei2024video} are often integrated with self-reflection or chain-of-thought to enhance their answering performance and interpretability.

The assessment of long video understanding requires multiple aspects including object recognition, temporal reasoning, and memory retention. Recent benchmark developments have primarily centered on question-answering (QA) tasks. These~\cite{cinepile, moviechat, movqa, vbench, longclip, hourvideo, lvos, mmbenchvideo, videovista} include evaluations of egocentric videos, online content, movies, TV shows, surveillance footage, and so on. Some benchmarks focus on specific aspects like temporal reasoning~\cite{mlvu} and plot comprehension~\cite{longvideobench}, while others emphasize fine-grained information retrieval from extended sequences~\cite{videoniah, videomme}. These diverse evaluation methods help assess both perceptual accuracy and reasoning capabilities in processing long-form content.

\paragraph{MLLMs for Specific Vision Problems.} While conventional MLLMs have demonstrated strong performance in tasks like visual captioning and question answering, they are challenged in handling fine-grained visual tasks such as segmentation and temporal grounding that require precise predictions. To address these limitations, researchers have developed two main approaches: 1) The pixel-to-sequence (P2S) methodology~\cite{chen2023shikra, Ren2023TimeChat, grounded-videollm,  hawkeye, mplugowl} enables MLLMs to generate textual predictions directly. These systems incorporate specialized components like time-aware encoders and sliding video processors to enhance temporal information understanding. 2) The pixel-to-embedding (P2E) approach~\cite{bai2024one, lai2024lisa, wang2024visionllm, wu2024visionllm, zhang2023nextchat} focuses on compressing visual information before passing it to specialized downstream decoders for final predictions. Recent implementations leverage advanced segmentation tools through prompt-based mechanisms, using special tokens to bridge MLLMs with segmentation modules. Some systems employ multiple routing tokens and enhanced query mechanisms to facilitate connections between MLLMs and various decoder components. These developments represent significant progress in creating more capable and versatile visual understanding systems.

\begin{figure*}[t]
    \centering
    \includegraphics[width=0.7\linewidth]{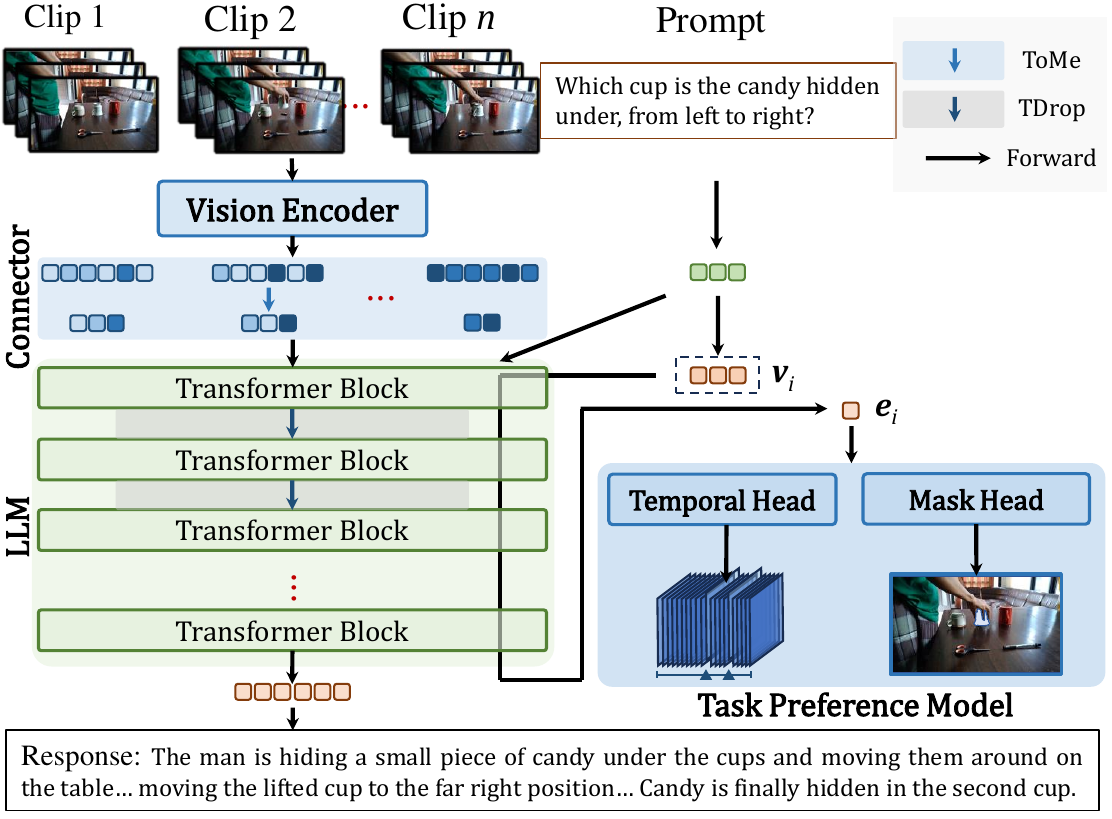}
    \caption{Framework of \modelname{} with the long and rich context (LRC) modeling.}
    \label{fig:overview}
\end{figure*}

\section{{\modelname}: Long and Rich Context Modeling}

To enable long and accurate video understanding with MLLM, we build {\modelname} to enhance the MLLM context length and fineness, employing a video length adaptive token representation and task preference optimization, as given in Figure \ref{fig:overview}. The whole model is learnt with three stages leveraging both short and long video data, as well as classical vision task data. The whole method is detailed below.

\subsection{Video Length Adaptive Token Representation For Long Multimodal Context}
Our approach introduces a practical length-adaptive token representation approach for processing video sequences of any lengths efficiently. This method is built upon a typical MLLM architecture comprising three key components: a vision encoder, a vision-language connector, and a language model. 
After a dynamic frame sampling, the given pipeline implements a hierarchical token compression (HiCo) with two distinctive stages: 1) spatiotemporal-aware compression during the visual encoding, and 2) adaptive multimodal context consolidation during language model processing.

\paragraph{Adaptive Temporal Sampling.}
We implement a context-aware sampling mechanism that adjusts according to video duration and content characteristics. For shorter sequences where motion granularity is crucial, we employ dense temporal sampling (15 frames per second). Conversely, for long sequences (e.g. minute / hour-level videos) focused on event-level understanding, we utilize sparse sampling (1 frame per second). This adaptive approach ensures proper motion capture across varying temporal scales.

\paragraph{Hierarchical Token Compression.} We compress long visual signals via their spatiotemporal redundancies in events and semantic redundancies between events.

\begin{itemize}[leftmargin=*]
    \item \textbf{Spatiotemporal Token Merging.} 
    To address the inherent temporal redundancy in video sequences, we process the input through a hierarchical compression scheme. Given a video sequence divided into $T$ temporal segments, each segment is processed by a vision encoder $\mathcal{E}$ to generate $M$ initial tokens: $[\mathbf{v}_i^j]_{i=1,2,..,M}$ for the $j$-th segment. These tokens undergo adaptive compression through a token connector $\mathcal{C}$, producing $N$ compressed tokens $[\mathbf{s}_i^j]_{i=1,2,..,N}$ where $N < M$:
    \begin{equation}
    [\mathbf{s}_i^j]_{i=1,2,..,N} = \mathcal{C}([\mathbf{v}_i^j]_{i=1,2,..,M}).
    \end{equation}
    To keep the essences of spatiotemporal tokens, token merging \cite{tome} can be treated as a semantic pooling operation, pooling tokens with high similarities instead of close locations. Such relation calculation is realized by a bipartite soft matching with a few of iterations. Prior empirical analysis ~\cite{li2024videochat} across various compression configurations (pooling, MLP, Q-Former, and more) reveal that exploiting semantic similarity-based token merging \cite{tome} (ToMe) across spatiotemporal scales as $\mathcal{C}$ shows the superiority in visual compression while preserving essential details. 
    Meanwhile, we argue query-conditioned visual token compression by Q-Former are no longer performant competitive and user-friendly in the context of up-to-date LLMs. A Q-Former is usually with 300M learanble parameters demanding several epochs of learning on at least 10M visual-text pairs. Such learning is independent of the pretraining or supervised finetuning (SFT) of vision or language models, while ToMe is plugable in training or testing without extra tuning just like pooling.

    \item \textbf{Multimodal Token Dropout.}
    We introduce the token dropout that operates during language model processing to further optimize long-range visual understanding. It implements a two-phase token reduction strategy: (1) uniform token pruning in early layers to maintain structural integrity while reducing computational overhead, and (2) attention-guided token selection in deeper layers to retain task-relevant essences.

    For $NT$ compressed tokens $[\mathbf{s}_i^j]$ from the video sequence, we apply this pruning during language model computation. Denoting $[\mathbf{t}_i^{(l)}]$ as the token representations at layer $l$, the pruning operation is defined as:
    \begin{equation}
    [\mathbf{t}_i^{(l)}] = [\mathbf{p}_i^{(l)} \odot \mathbf{t}_i^{(l-1)}] \quad \text{where} \quad \mathbf{p}_i^{(l)} \sim \text{Bernoulli}(p),
    \end{equation}
    where $p$ denotes the token preservation probability.
    This adaptive pruning mechanism not only enhances computational efficiency but also improves model performance by reducing irrelevant visual information in the token representation.
\end{itemize}

Note all three steps in the given video length adaptive token representation are non-learnable. Yet, training MLLMs with HiCo would yield much better performance as HiCo benefits from training in more context-adaptive features.

\subsection{Enhancing Visual Precision in Multimodal Context through Task Preference Optimization}
We enhance multimodal language models' (MLLMs) precise visual understanding capabilities through Multi-task Preference Learning (MPL). Our approach integrates specialized visual perception modules with a base MLLM architecture to enable fine-grained visual analysis capabilities such as precise localization and temporal understanding. The framework consists of a core MLLM ($\Phi$) containing a visual encoder ($\mathcal{E}$), cross-modal connector ($\mathcal{C}$), and language model ($\mathcal{L}$), augmented with a specialized visual perception module ($\mathcal{P}$) comprising task-specific heads ${H_k}$ (where $k=1,2,..., K$). These heads interface with the MLLM through learned task embeddings ${\mathbf{e}_k}$ derived from trainable task tokens ${\mathbf{v}_k}$.

\paragraph{Visual Perception Preference.}
The visual perception module incorporates two fundamental capabilities essential for precise visual understanding:

\begin{itemize}[leftmargin=*]

    \item \textbf{Temporal Understanding.} For processing dynamic visual content, we develop a temporal component that combines video feature extraction with temporal alignment capabilities. The module takes in both visual sequences and textual queries, incorporating task-specific temporal embeddings to predict precise temporal boundaries and relevance scores.

    \item \textbf{Instance Segmentation.} To enable pixel-precise understanding and instance-level differentiation, we design a segmentation module building upon recent advances in foundation models for segmentation. The module consists of an image encoder, mask decoder, and an adaptive projection layer that bridges MLLM embeddings with pixel-level predictions.
\end{itemize}

To effectively integrate these specialized capabilities while preserving the MLLM's general capabilities, we optimize the MLLM $\Phi$ along with visual perception module $\mathcal{P}$ as:
\begin{equation}
\mathcal{L}_{\text{total}} = \mathcal{L}_{\text{base}} + \lambda_1\mathcal{L}_{\text{task}}(G(\mathbf{Q}), \mathbf{y}) + \lambda_2\sum_{k=1}^K \mathcal{L}_{\text{spec}}(\mathbf{Y}_k, H_k(G(\mathbf{v}_k))),
\end{equation}
where $\mathbf{Q}$ represents input queries, $\mathbf{y}$ denotes task labels, and $\mathbf{Y}_k$ represents task-specific annotations.

Our framework improves the MLLM's general capabilities by significantly enhancing its precision in specific visual analysis tasks. The modular design allows for flexible extension to additional visual understanding capabilities while maintaining efficient computation and training dynamics.

\subsection{Training Video Corpus for Multimodal Context Modeling} 

The training process has three stages, in which visual-text aligned data, long video data and task-specific visual data are used. The training data can be seen in Table~\ref{atab:data}.

\paragraph{Visual-Text Data For Crossmodal Alignment.} We curate a collection of visual-text data with 7M image-text pairs and 3.7M video-text pairs, as well as 143K text data for language capability enhancement. For visual-caption pairs, we convert them into a question-answering (QA) format for training convenience. Specifically,
\begin{itemize}[leftmargin=*]
    \item \textbf{Image-Text.} We utilize 3.5 million detailed image descriptions recaptioned with LLava-NeXT-34B~\cite{llavanextvideo} from COCO118K, BLIP558K, and CC3M datasets~\cite{llavaonevision}. For instruction data, we use single-image instructions from LLava-NeXT~\cite{llavanextvideo}, Allava~\cite{allava}, and ShareGPT4O~\cite{internvl2,internvideo2}, along with multi-image instructions from LLaVA-Interleave~\cite{llavainterleave}. Additionally, we incorporate 558K image-text pairs from LCS-558K~\cite{llava}.

    \item \textbf{Video-Text.} Our video-text data consists of WebVid2M~\cite{webvid} descriptions recaptioned with VideoChat2~\cite{mvbench}, and 323K detailed descriptions from WebVid~\cite{webvid} and Kinetics~\cite{kinetics} recaptioned with Gemini~\cite{gemini}\cite{sharegemini}. For instruction fine-tuning, we use short video data from VideoChat2\cite{mvbench} and InternVideo2~\cite{internvideo2}, complemented by GPT4o-annotated data from ShareGPT4o~\cite{internvl2,internvideo2}, VideoChatGPT-Plus~\cite{videogpt+}, LLaVA-Video-178K~\cite{llavavideo}, and LLava-Hound~\cite{llavahound}.
    
    \item \textbf{Text.} We incorporate 143K samples from the Evo-Instruct dataset~\cite{allava}.
\end{itemize}

\paragraph{Long Video Corpus for Context Extension.} We primarily utilized long video instruction data from MoiveChat~\cite{moviechat}, Cineplie~\cite{cinepile}, Vript~\cite{vript} and our LongVid. A significant challenge in training long video models is the scarcity of large-scale, high-quality data. Although recent progress has alleviated this issue to some extent through long-form datasets of video-text pairs, these datasets lack the instruction-following paradigm, such as (video, instruction, answer) triplets, which is essential for multimodal reasoning. To tackle this problem, we employ a large-scale long video instruction-tuning dataset named LongVid from \cite{li2024videochat}. This dataset consists of 114,228 long videos and 3,444,849 question-answering (QA) pairs across five distinct task types, enabling models to handle various long video scenarios.

\begin{table*}[ht]
\centering
\resizebox{1.\linewidth}{!}{
    \begin{tabular}{l|c|c|c}
    \toprule
    \textbf{Stage} & \textbf{Task} &\textbf{Samples} & \textbf{Datasets}\\
    \midrule 
    \multirow{4}{*}{Stage 1} 
    & Segmentation & 50K & SAMv2, MeViS \\
    & Temporal Grounding & 50K &  DiDeMo, QuerYD \\
    & Spatial Grounding & 50K & RefCOCO, RefCOCOg, RefCOCO+ \\
    & Alignment data & 1M & LCS-558K, S-MiT \\
    \midrule 
    \multirow{4}{*}{Stage 2} 
    & Segmentation & 114.6K & SAMv2, MeViS \\    
    & Temporal Grounding & 116.5K & DiDeMo, QuerYD, HiRest, ActivityNet, TACoS, NLQ  \\
    & Spatial Grounding & 540.0K & AS-V2, Visual Genome, RefCOCO, RefCOCO+, RefCOCOg \\
    & Visual Concept & 6M & VideoChat2-IT, WebVid2M, Share-Gemini, LLaVA-NexT, Evo-Instruct\\
    \midrule 
    \multirow{5}{*}{Stage 3} 
    & Temporal Grounding  & 7.5K & QVHighlight \\
    & Segmentation & 116.5K & MeViS, SAMv2 \\
    & Temporal Reasoning & 40K & YouCook2, ActivityNet \\
    & Spatial Grounding & 400K & AS-V2, Visual Genome, RefCOCO, RefCOCO+, RefCOCOg \\
    & \multirow{2}{*}{Conversation}  & \multirow{2}{*}{3.5M} & LLaVA-NexT, Evo-Instruct,ShareGPT4o,LLaVA-Interleave, VideoChat2-IT,\\
    & & &  VideoChatGPT-Plus, LLaVA-Video, LLaVA-Hound, MovieChat, Vript, LongVid\\
\bottomrule 
\end{tabular}
}
\caption{The training data specifications encompass more than the standard public video question-answering datasets. In addition to common data, we incorporate annotations from lengthy videos and typical vision task data.}
\label{atab:data}
\end{table*}

\paragraph{Task-Specific Data for Accurate Perception.} 
\begin{itemize}[leftmargin=*]
    \item \textbf{Segmentation.}  We use MeViS~\cite{mevis} and SAMv2~\cite{ravi2024sam2} for referring segmentation task.

    \item \textbf{Spatial Grounding.} We use AS-V2~\cite{wang2024allseeing_v2}, Visual Genome~\cite{krishna2017vg}, RefCOCO~\cite{yu2016refcoco}, RefCOCOg~\cite{yu2016refcoco}, RefCOCO+~\cite{yu2016refcoco} for one epoch with a total batch size of 128 to train region head and token.

    \item \textbf{Temporal Grounding.} We utilize DiDeMo~\cite{DiDeMo}, QuerYD~\cite{queryd}, HiRest~\cite{hirest}, ActivityNet~\cite{caba2015activitynet}, TACoS~\cite{regneri2013grounding}, NLQ~\cite{grauman2022ego4d}.

\end{itemize}

\subsection{Progressive Multi-stage Training}
We propose a unified progressive training scheme that jointly enhances MLLM's fine-grained perception and temporal understanding abilities. Our approach consists of three main stages that gradually increase both the complexity of tasks and the temporal length of video inputs.

\paragraph{Stage 1: Foundation Learning.} This initial stage focuses on two parallel objectives: (a) task recognition instruction tuning for LLM using diverse dialogue templates, enabling the model to identify and route different visual tasks; and (b) video-language alignment training where we freeze the visual encoder and LLM while optimizing the compressor and MLP to establish basic visual-linguistic connections. We utilize 0.5M image-text pairs and 0.5M short video-text pairs with 4 frames per video in this stage.

\paragraph{Stage 2: Fine-grained Perception Training.} This stage enhances the model's visual understanding capabilities through: (a) integration and training of task-specific components including task tokens, region head, temporal head, and mask adapter using task-specific datasets; and (b) visual concept pre-training using 3.5M images and 2.5M short video-text pairs with 8 frames per video. The LLM is updated using LoRA during this stage to maintain its general capabilities while acquiring new visual skills.

\paragraph{Stage 3: Integrated Accurate and Long-form Context Training.} The final stage jointly optimizes all model components through: (a) multi-task training on mixed corpus combining multimodal conversations and specific task data, allowing task-supervised gradients to flow from specialized heads to MLLM; and (b) instruction tuning on a comprehensive dataset of 3.5M samples, including 1.1M images, 1.7M short videos (<60s), and 0.7M long videos (60-3600s). We employ dynamic video sampling with 64-512 frames and tune the entire model including the vision encoder, connector, task tokens, specialized heads, and LLM with LoRA.

This progressive training strategy enables the model to develop both fine-grained perception and long-form video understanding while mitigating potential degradation of general abilities. Unlike previous approaches that rely on long-form text for extending context windows, our direct training on long-form videos minimizes the gap between training and deployment scenarios.

\subsection{Implementation}

\paragraph{Distributed System.}
We develop a multimodal sequence parallelism system based on XTuner to train and test on millions of multimodal tokens (mostly in visual). Built upon several open-sourced solutions \cite{longvila,usp,deepspeed-ulysses, deepspeed,ringattention}, we enable scalable computing of long videos by integrating both sequence and tensor distributed processing as well as multimodal dynamic (soft) data packing. 

To efficiently distribute multimodal token sequence, we divide it in head processing in multi-head self-attention (MHSA) and input sequence length. Specifically, we employ All-to-All (A2A) communication for tensor parallel computing from DeepSpeed-Ulysses \cite{deepspeed-ulysses}. As it is limited to head number in attention, we further incorporate ring-attention \cite{ringattention} to realize concurrent computing on the sequence for rising parallelism degree in Peer-to-Peer (P2P) communication. Regarding high transfer speed in inter-node and low speed in intra-node settings, we utilize the 2D-attention strategy \cite{usp} that assigns A2A to inter-node computing and P2P to intra-node, as any device is required to communicate with any others for A2A while any device is with the two neighbors in the ring structure for P2P.

In regard to the variance in training video length, we need unify the data size for batch training by packing them (or sharding them in some studies). A simple padding strategy widely adopted in mainstream MLLM learning is padding the short sequence and clip the long one to the target length in the same batch. Despite its simplicity, its data utilization effectency has room to improve as it fills too many placeholders in inputs. In contrast, we employ a dynamic packing strategy. Given a fixed sequence length in each training iteration, it dynamically merge input sequences by order into a new one (ensuring its total length do not exceed the target). Note it can maximize the GPU memory usage to achieve speedup ratio, especially when the distribution of training video lengths is beyond even.

\paragraph{Model Configuration.} In our multimodal architecture, we utilize a comprehensive framework that combines advanced video processing and language modeling capabilities. The system implements dynamic video sampling, processing 64-512 frames, with each 8-frame clip compressed to 128 tokens, yielding approximately 16 tokens per frame representation. The architecture integrates InternViT \cite{chen2024expanding} for visual encoding, along with an MLP-based token merging mechanism and InternLM2.5-7B \cite{cai2024internlm2} as the language model.

The framework incorporates multiple specialized components: a Temporal Head based on CG-DETR architecture \cite{moon2023cgdetr}, and a Mask Head utilizing SAM2's pre-trained weights. For temporal processing, the system leverages InternVideo2 \cite{internvideo2} for video feature extraction, while query features are processed through the language model. To enhance spatiotemporal capabilities, we implement two-layer MLPs for both positioning prompts and spatial input encoding into the multimodal language model.

\section{Experiments}

We evaluate InternVideo2.5-7B along with its base model and other comparisons on mainstream multimodal video understanding and classical vision tasks. With the incorporation of HiCo and TPO to MLLMs, InternVideo2.5 can address short and long video dialogue (in Table \ref{tab:main}) and typical vision tasks like tracking (in Table \ref{tab:specific_vision_tasks}).

\begin{table*}[t]
    \centering
\begin{adjustbox}{width=\linewidth,center}
\renewcommand{\arraystretch}{1.1}
\setlength{\tabcolsep}{1.5mm}
\begin{tabular}{lrcccccccccc}
\toprule  {\textbf{Model}} & \multicolumn{1}{c}{{\centering \textbf{Size}}} & {\textbf{\#Tokens}}  & {\textbf{MVBench}} & {\textbf{PerceptionTest } } & {\centering \textbf{EgoSchema} }  &{\textbf{LongVideoBench}} & {\textbf{MLVU}} & {\centering \textbf{VideoMME}}  & {\textbf{LVBench}}  \\
Average duration (sec) & &  & 16 & 23 & 180 & 473 & 651 &  1010 & 4101\\
\midrule
\textit{Proprietary Models} \\
GPT4-V~\citep{gpt4v} & - & - & 43.7 & - & - & 59.1  & 49.2 & 59.9 & -\\
GPT4-o~\citep{gpt4o} & - & - & 64.6 & - & 72.2 & 66.7  & 64.6 & 71.9  & 30.8\\
Gemini-1.5-Pro~\citep{gemini} & - & - & 60.5 & - & 71.2  & 64.0  & - & 75.0  & 33.1 \\
\midrule
\textit{Open-Source MLLMs} \\
InternVL2~\citep{internvl2} & 8B & 256 & 66.4 & - & - & -  & - & 54.0 & -\\
\color{gray}InternVL2~\citep{internvl2} & \color{gray}76B & \color{gray}256 & \color{gray}69.6 & \color{gray}- & - & \color{gray}-  & \color{gray}- & \color{gray}61.2 & -\\
LLaVA-NeXT-Video~\citep{llavanextvideo} & 7B & 144 & 53.1 & 48.8 & - & 49.1 & - & 46.5  & -\\
LLaVA-OneVision~\citep{llavaonevision} & 7B & 196 &56.7 & 57.1  & 60.1 & 56.3  & 64.7 & 58.2  & -\\
\color{gray} LLaVA-OneVision~\citep{llavaonevision} & \color{gray} 72B & \color{gray}196 &\color{gray}59.4 & \color{gray}66.9  & - & \color{gray}61.3  & \color{gray}68.0 & \color{gray}66.2 & \color{gray}26.9\\
VideoLLaMA2~\citep{videollama2} & 7B & 72 & 54.6 & 51.4 & 51.7 & -  & 48.5 & 47.9 & -\\
\color{gray}VideoLLaMA2~\citep{videollama2} & \color{gray} 72B &\color{gray} 72 & \color{gray}62.0 & \color{gray}57.5 & \color{gray}63.9 & \color{gray}-  &\color{gray}- & \color{gray}62.4  & -\\
mPLUG-Owl3~\citep{ye2024mplug} & 7B & - & 54.5 & -  & - & 52.1  & - & 53.5 & 43.5\\
QwenVL2~\citep{qwen} & 7B & - & 67.0 & 62.3  & 66.7 & -  & - & 63.3 & -\\
\color{gray}QwenVL2~\citep{qwen} & \color{gray}72B & \color{gray}- & \color{gray}73.6 & \color{gray}68.0  & \color{gray}77.9 & \color{gray}-  & \color{gray}- & \color{gray}71.2 & \color{gray}41.3\\
VideoChat2-HD~\citep{mvbench} & 7B & 72 & 62.3 & - & - & -  & 47.9 & 45.3 & -\\
InternVideo2-HD~\citep{internvideo2} & 7B & 72 & 67.2 & 63.4  & 60.0 & -  & - & 49.4 & -\\
VideoChat-TPO \cite{yan2024task} & 7B & 64 & 66.8 & - & - & -  & 54.7 & - & -\\
\rowcolor{lightgray!20} InternVL2.5 \cite{chen2024expanding} & 7B & 256 & 72.0 & 68.2 & 51.5 & 60.0  & 68.9 & 64.2 & 38.4\\
\midrule
\textit{Open-Source Long Video MLLMs} \\

LLaMA-VID~\citep{llamavid} & 7B & 2 & 41.9 & 44.6 & - & - & 33.2 & 25.9 & 23.9\\
LongVILA~\citep{longvila} & 7B & 196 & - & - & 67.7 & -  & - & 57.5 & -\\
LongVA~\citep{longva} & 7B & 144 & - & - & - & -  & 56.3 & 52.6 & -\\
LongLLaVA~\citep{longllava} & 9B & 144 & 49.1 & - & - & -  & -& 43.7 & -\\
LongVU~\citep{longvu} & 7B & 64 & 66.9 & - & 67.6 & -  & 65.4 & - & -\\
VideoChat-Flash \cite{li2024videochat} & 7B & 16 & 73.2 & 75.6 & - & 64.2  & 74.5 & 64.0 & 47.2\\
\rowcolor{lightgray!20} \textbf{\modelname~(InternVL2.5+LRC)} & 7B & 16 & 75.7$_{(+3.7)}$ & 74.9$_{(+6.7)}$ & 63.9$_{(+12.4)}$ & 60.6$_{(+0.6)}$  & 72.8$_{(+3.9)}$ & 65.1$_{(+0.9)}$  & 46.4$_{(+8.0)}$ \\
\bottomrule
\end{tabular}
\end{adjustbox}
\caption{Performance on video question-answering benchmarks, covering both short and long videos. The models and their corresponding results, displayed in gray font, are based on LLMs of over 7 billion parameters in size.
}
\label{tab:main}
\end{table*}

\subsection{Video Understanding} 
As given in Table \ref{tab:main}, {\modelname} achieves almost the leading performance in all popular short and long video question-answering benchmarks in around 7B LLM capacity level. Compared with the employed base MLLM InternVL2.5, {\modelname} gives a overall increase no matter on short or long duration predictions. The rises are notable when dealing with short videos, {\modelname} increases more than 3 points on MVBench and Perception Test based on InternVL2.5. Concerning long video understanding, the whole trend is still rising but the variations are changing on different benchmarks. The brought increases are clear on EgoSchema (test) (+12.4) \cite{egoschema}, MLVU (+3.9) \cite{mlvu}, and LVBench (+8.0) \cite{wang2024lvbench} while they are relatively marginal on LongVideoBench (+0.6) \cite{longvideobench} and VideoMME (wo sub) (+0.9) \cite{videomme}. We attribute this to the questions in the latter two benchmarks are more dependent of world knowledge and reasoning rather than perceived evidences, since the performance increases over them are more apparent with the larger language models.

In the comparison with renown proprietary models e.g. GPT4-o and Gemini-1.5-Pro, {\modelname} shows superiority over short duration spatiotemporal understanding while gives inferior results on long videos (except on MLVU). This may imply that we still have much room to explore for the vision-language fusion and long context modeling though we do make a huge step in vision areas from the academic and open-sourced perspective.

\begin{figure*}[t]
    \centering
    \includegraphics[width=0.8\linewidth]{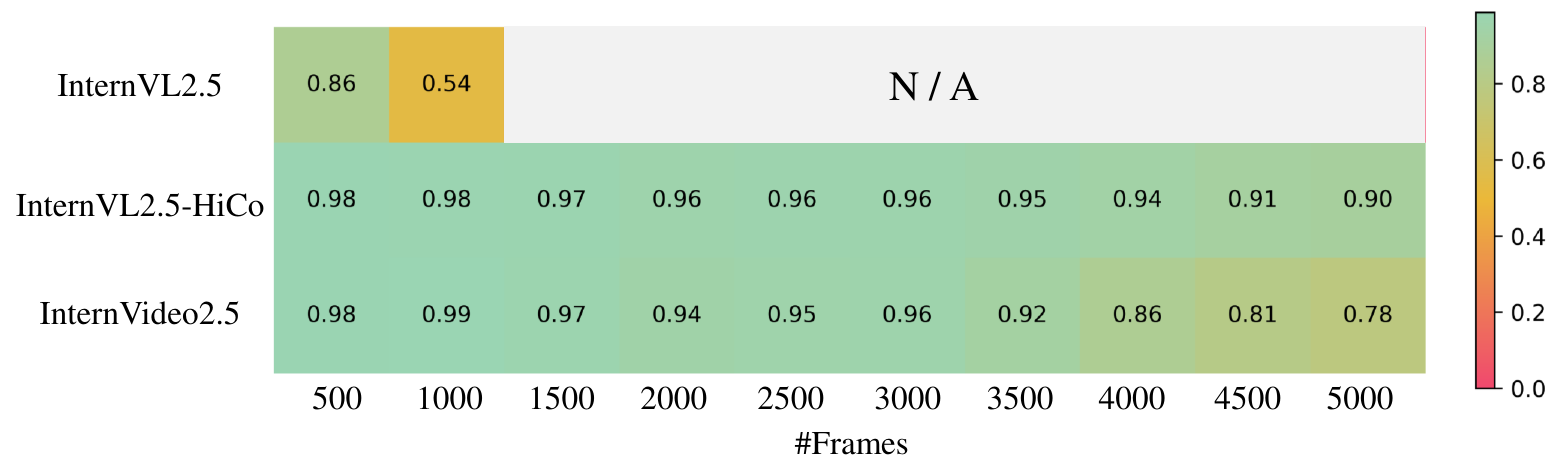}
    \caption{Single-hop Needle-in-a-haystack evaluation results using 5,000 frames using 16 A100 (80G) GPUs.}
    \label{fig:niah}
\end{figure*}

\paragraph{Needle-In-The-Haystack (NIAH).} With HiCo and long video learning corpus, InternVideo2.5 notably enhances InternVL2.5's implicit memory (8B model). Using a single-hop needle-in-a-haystack (NiAH) task with 5,000 frames, InternVideo2.5 demonstrates superior recall. All models were evaluated on 16 A100 (80GB) GPUs.

As shown in Figure \ref{fig:niah}, InternVL2.5-8B struggles to recall target frames accurately even within 500 frames, falling short of the 95\% accuracy threshold for memorization established in \cite{longva,longvila}. Furthermore, processing long video inputs leads to its out-of-memory (OOM) errors beyond 1,000 frames. In contrast, InternVideo2.5, benefiting from HiCo and long video training, accurately recalls frames from up to 3,000 frame sequences and processes over 10,000 frames without OOM issues. While the LRC variant (HiCo + TPO) maintains high recall up to 3,000 frames, its performance degrades beyond this point, likely due to the discrepancy between its training data (primarily short videos) and the long video evaluation setting. Future work could explore adjusting the data ratio and incorporating long-form video annotations for TPO to mitigate this issue. Additionally, the context modeling of the LLM directly impact its MLLM's context. Therefore, enhancing the LLM's context or utilizing a more effective one could further improve performance.

\begin{figure*}[t]
    \centering
    \includegraphics[width=0.8\linewidth]{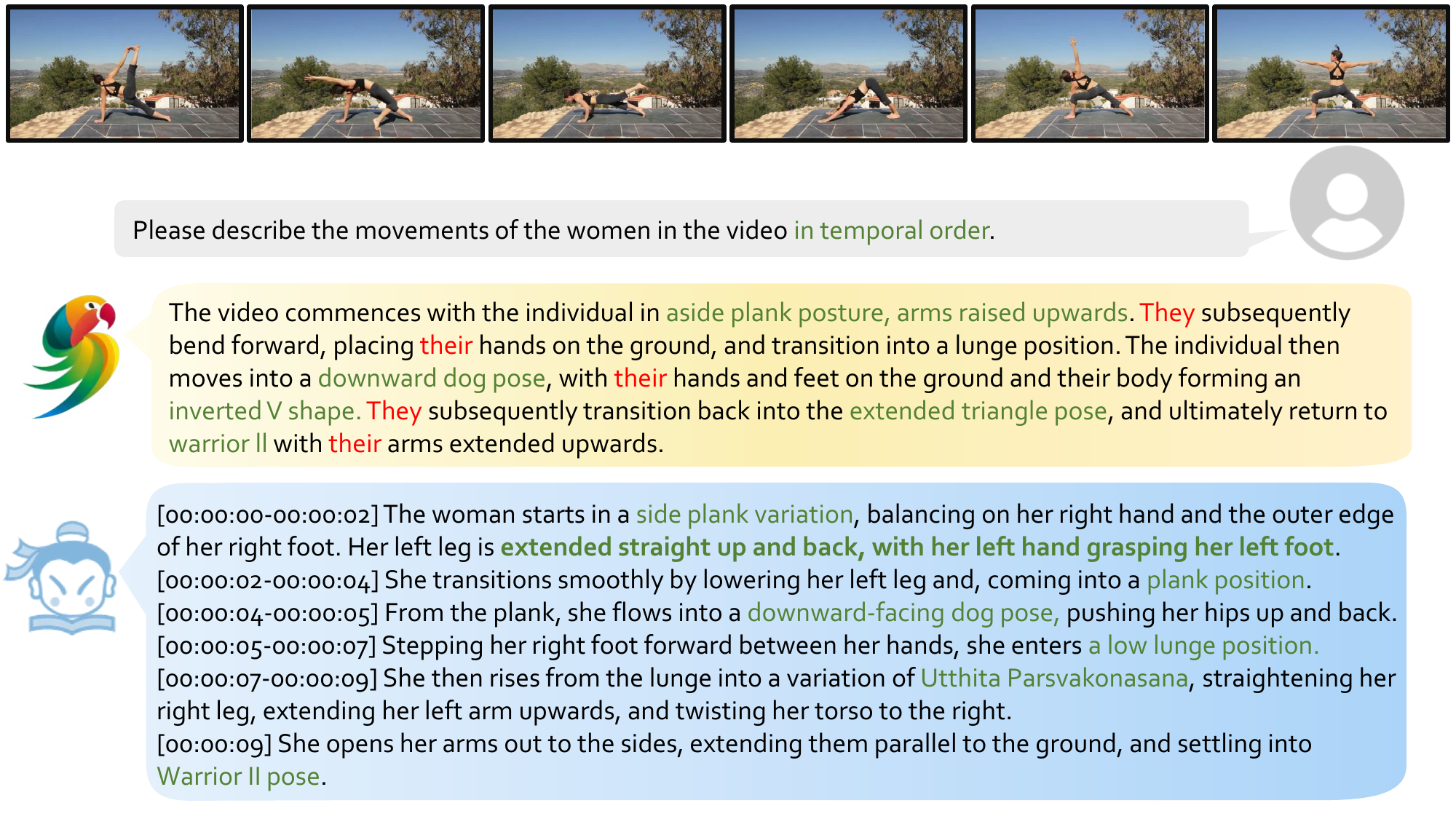}
    \caption{An example of {\modelname}: it can depict spatiotemporal movements with precise temporal references.}
    \label{fig:c1}
\end{figure*}

\begin{figure*}[t]
    \centering
    \includegraphics[width=0.8\linewidth]{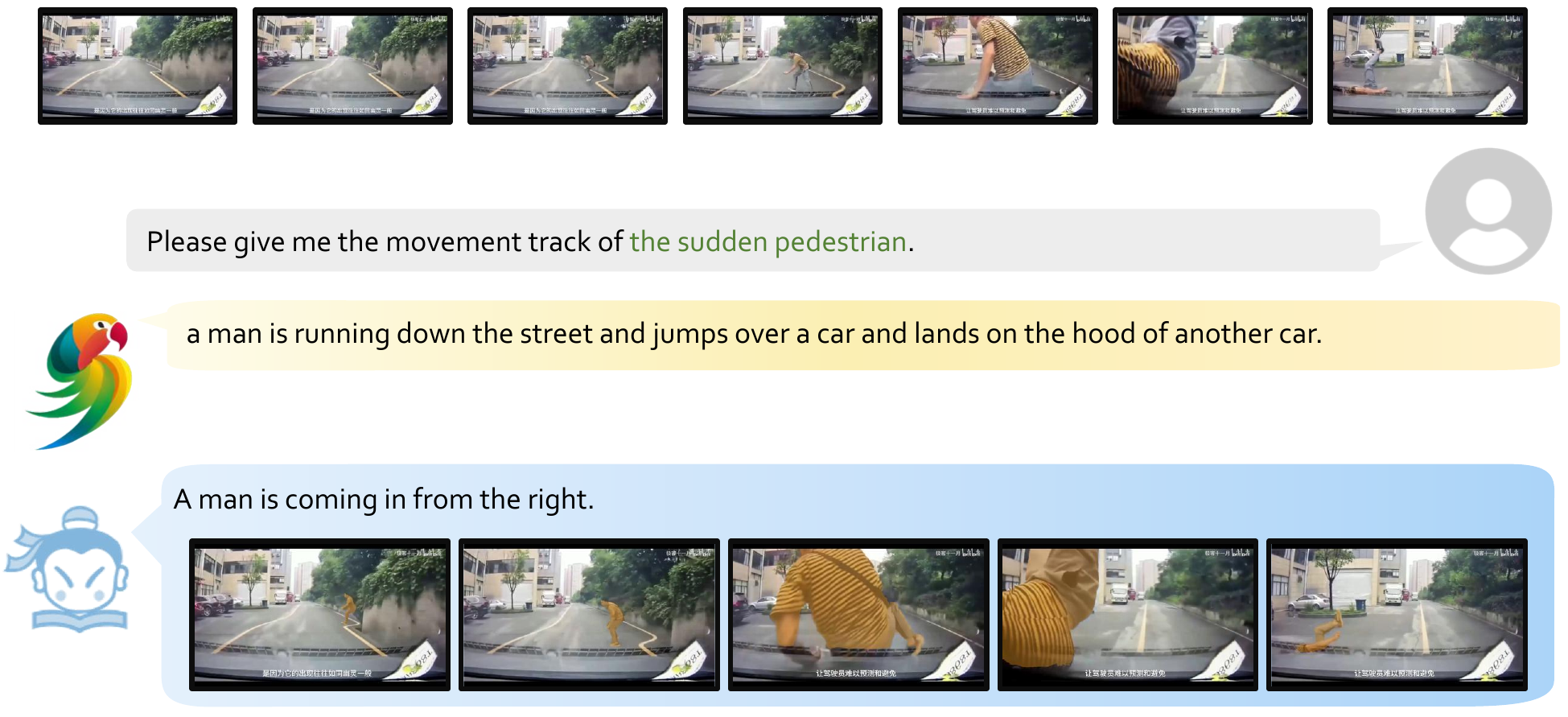}
    \caption{An example of {\modelname}: it can track the target specified by users and reason over it.}
    \label{fig:c3}
\end{figure*}

\begin{figure*}[t]
    \centering
    \includegraphics[width=0.8\linewidth]{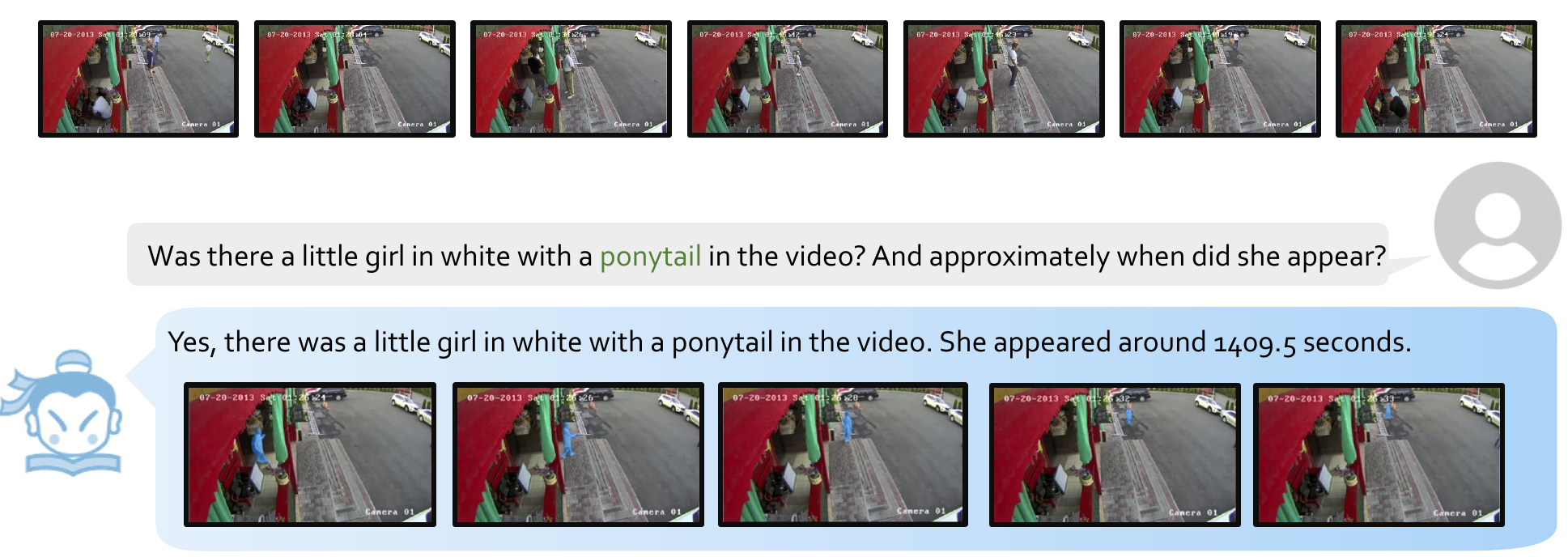}
    \includegraphics[width=0.8\linewidth]{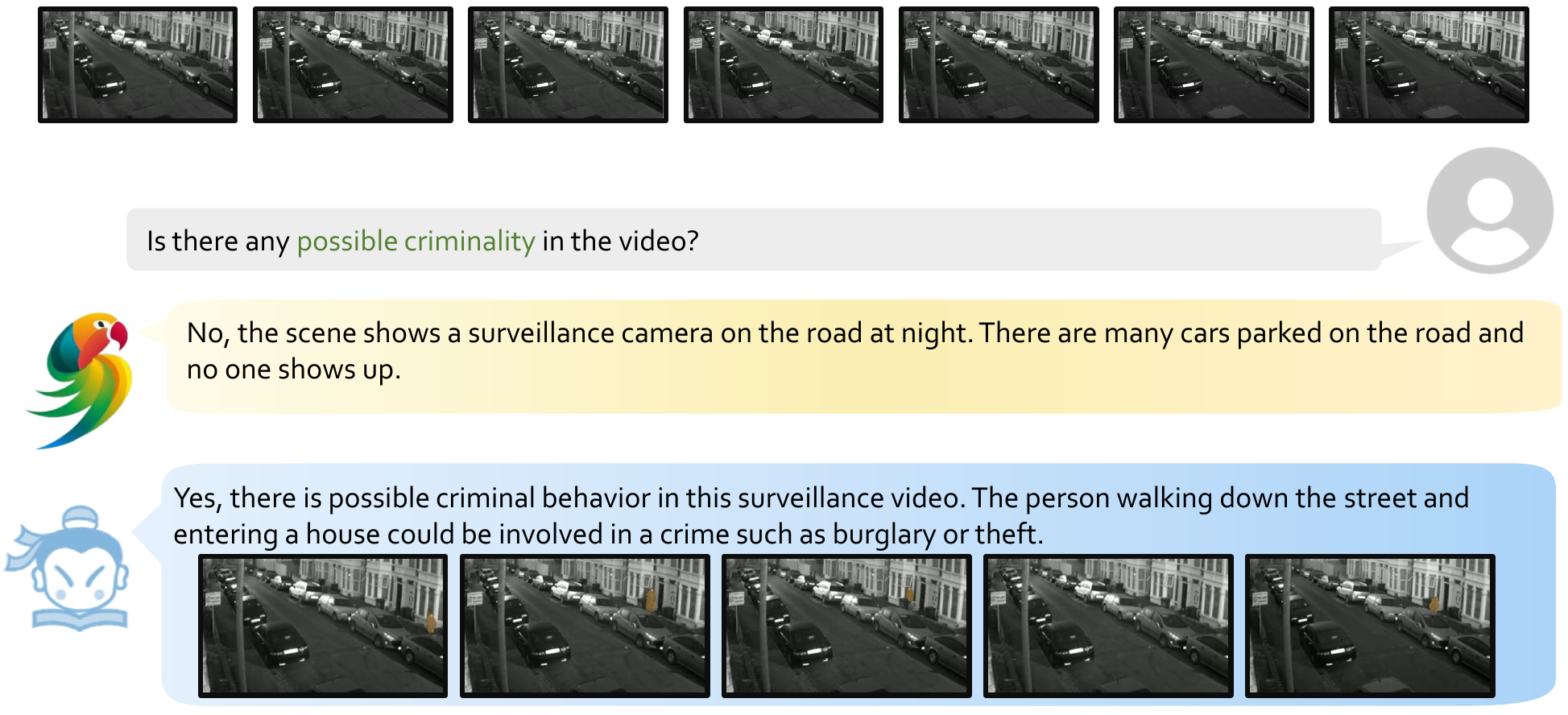}
    \caption{Examples of {\modelname}: it can can comprehend extended surveillance videos for moment retrieval or abnormal event detection.}
    \label{fig:c2}
\end{figure*}

\paragraph{Qualitative Evaluations.} We give several real cases evaluated by {\modelname} and VideoChat2 \cite{mvbench} in Figure \ref{fig:c1}, \ref{fig:c3}, and \ref{fig:c2}. Note {\modelname} can give detailed motion descriptions as well as their specific times, track (or segment) the target for further reasoning, and understand over a long visual input.

\begin{table*}[t]
\centering
    \resizebox{\textwidth}{!}{
    \begin{tabular}{lcccccccccc}
    \toprule
    \multirow{2}{*}{Method} & \multicolumn{2}{c}{Charades-STA} & \multicolumn{2}{c}{Highlight Detection} & Ref-YouTube-VOS & MeViS & \multicolumn{2}{c}{LaSOT} & \multicolumn{2}{c}{GOT-10k} \\ 
     & R@0.5 & mIoU & mAP & HIT@1 & J\&F & J\&F & Success & P$_\text{norm}$ & Overlap & SR$_{0.5}$ \\ \hline
    \textcolor{gray}{UniVTG~\cite{univtg}} & \textcolor{gray}{25.2} & \textcolor{gray}{27.1} & \textcolor{gray}{40.5} & \textcolor{gray}{66.3} & \textcolor{gray}{-} & \textcolor{gray}{-} & \textcolor{gray}{-} & \textcolor{gray}{-} & \textcolor{gray}{-} & \textcolor{gray}{-} \\ 
    \textcolor{gray}{ReferFormer~\cite{wu2022language}} & \textcolor{gray}{-} & \textcolor{gray}{-} & \textcolor{gray}{-} & \textcolor{gray}{-} &  \textcolor{gray}{62.9} &  \textcolor{gray}{31.0} & \textcolor{gray}{-} & \textcolor{gray}{-} & \textcolor{gray}{-} & \textcolor{gray}{-} \\ 
    \textcolor{gray}{OnlineRefer~\cite{wu2023onlinerefer}} & \textcolor{gray}{-} & \textcolor{gray}{-} & \textcolor{gray}{-} & \textcolor{gray}{-} &  \textcolor{gray}{62.9} & \textcolor{gray}{-} & \textcolor{gray}{-} & \textcolor{gray}{-} & \textcolor{gray}{-} & \textcolor{gray}{-} \\ 
    \textcolor{gray}{TFVTG(GPT-4T)~\cite{zheng2025training}} & \textcolor{gray}{49.9} & \textcolor{gray}{44.5} & \textcolor{gray}{-} & \textcolor{gray}{-} & - & - & - & - & - & - \\ 
    VideoChat2~\cite{mvbench} & 14.3 & 24.6 & - & - & - & - & - & - & - & - \\ 
    VTimeLLM~\cite{vtimellm} & 27.5 & 31.2 & - & - & - & - & - & - & - & - \\ 
    TimeChat~\cite{Ren2023TimeChat} & 32.2 & - & 21.7 & 37.9 & - & - & - & - & - & - \\ 
    HawkEye~\cite{hawkeye} & 31.4 & 33.7 & - & - & - & - & - & - & - & - \\ 
    ChatVTG~\cite{chatvtg} & 33.0 & 34.9 & - & - & - & - & - & - & - & - \\ 
    LISA~\cite{lai2024lisa} & - & - & - & - & 52.6 & - & - & - & - & - \\ 
    VideoLISA~\cite{bai2024one} & - & - & - & - & 63.7 & 44.4 & - & - & - & - \\ 
    SiamFC~\cite{siamFC} & - & - & - & - & - & - & 33.6 & 42.0 & 34.8 & 35.3 \\ 
    ATOM~\cite{ATOM} & - & - & - & - & - & - & 51.5 & - & 55.6 & 63.4 \\ SiamRPN++~\cite{SiamRPN++} & - & - & - & - & - & - & 49.6 & 56.9 & 51.8 & 61.8 \\ 
    SiamFC++~\cite{SiamFC++} & - & - & - & - & - & - & 54.4 & 62.3 & 59.5 & 69.5 \\ 
    LLaVA-1.5~\cite{llava} & - & - & - & - & - & - & 19.4 & 16.5 & 23.5 & 20.2 \\ 
    Merlin~\cite{yu2025merlin} & - & - & - & - & - & - & 39.8 & 40.2 & 51.4 & 55.9 \\ \hline 
    VideoChat-TPO & \underline{40.2} & \underline{38.1} & \textbf{38.8} & \textbf{66.2} & \textbf{63.9} & \textbf{47.0} & \underline{69.4} & \underline{80.1} & \underline{70.6} & \underline{79.8} \\ 
    {\modelname} & \textbf{43.3} & \textbf{41.7} & \underline{34.7} & \underline{60.3} & 34.2 & 32.0 & \textbf{71.5} & \textbf{82.1} & \textbf{72.4} & \textbf{83.0} \\ \hline 
    \end{tabular}
    }
    \caption{Performance on Specific Visual Tasks. The grey means is an expert model without LLM, and the tasks they can handle are limited to those that are fine-tuned.}
    \label{tab:specific_vision_tasks}
    \end{table*}

\subsection{Specific Vision Tasks}

After the optimization of TPO, MLLMs can not only improve the understanding ability of video conversations, but also have the ability to handle specific densely annotated visual tasks, such as moment retrieval, reference tracking, and so on.
With TPO, MLLMs master classical vision capabilities e.g. tracking. Table~\ref{tab:specific_vision_tasks} presents that {\modelname} can conduct track, video referring segmentation, temporal grounding, and other tasks with expert model-level performance, clearly outperforming other MLLMs in most tasks. It also validates that the joint learning between mutimodal question-answering and typical vision tasks could benefit each other as in \cite{yan2024task}.

\subsection{Ablation Studies}

\begin{table*}[t]
    \centering
\begin{adjustbox}{width=.9\linewidth,center}
\renewcommand{\arraystretch}{1.1}
\setlength{\tabcolsep}{1.5mm}
\begin{tabular}{lrccccccccc}
\toprule  {\textbf{Base Model}} & {{\centering \textbf{\#Tokens}}}  & {\textbf{MVBench}} & {\textbf{PerceptionTest } } & {\centering \textbf{EgoSchema} }  & {\textbf{LongVideoBench}} & {\textbf{MLVU}} & {\centering \textbf{VideoMME} }  \\
\midrule
InternVL2.5-HiCo & 64 &  74.4 & 71.9 & 65.7 & 62.7 & 72.6 & 66.4 \\
InternVL2.5-HiCo & 16 & 74.0 & 71.4 & 62.9 & 59.6  & 71.5 & 64.9 \\
\bottomrule
\end{tabular}
\end{adjustbox}
\caption{Performance of InternVL2.5 using HiCo with varying numbers of tokens per frame.
}
\label{tab:hico_len}
\end{table*}

\paragraph{How token representation length affects Model Performance.} Table \ref{tab:hico_len} gives how token number for each frame in HiCo affects InternVL2.5 on video benchmarks. Note the decreases from fewer token number in HiCo on short video benchmarks are relatively marginal (around 0.5 points) while these are non-trivial (around 1-3 points) on long ones. This empirically verifies that there is still significant performance headroom to explore in long multimodal context.

\begin{table*}[!h]
    \centering
\begin{adjustbox}{width=.9\linewidth,center}
\renewcommand{\arraystretch}{1.1}
\setlength{\tabcolsep}{1.5mm}
\begin{tabular}{lrccccccccc}
\toprule  \multirow{2}{*}{\textbf{Base Model}} & \multicolumn{2}{c}{{\centering \textbf{LRC}}}  & \multirow{2}{*}{\textbf{MVBench}} & \multirow{2}{*}{\textbf{PerceptionTest } } & \multirow{2}{*}{\centering \textbf{EgoSchema} }  &\multirow{2}{*}{\textbf{LongVideoBench}} & \multirow{2}{*}{\textbf{MLVU}} & \multirow{2}{*}{\centering \textbf{VideoMME} }  \\
 & {\textbf{HiCo}}  & {\textbf{TPO}} &  &  & & & \\
\midrule
InternVL2.5 &  & & 72.0 & 68.2 & 51.5 & 60.0  & 68.9 & 64.2 \\
InternVL2.5 & \checkmark &  & 74.0 & 71.4 & 62.9 & 59.6  & 71.5 & 64.9 \\
InternVL2.5 & \checkmark & \checkmark & \textbf{75.7} & \textbf{74.9} & \textbf{63.9} & \textbf{60.6}  & \textbf{72.8} & \textbf{65.1} \\
\bottomrule
\end{tabular}
\end{adjustbox}
\caption{Evaluating HiCo and TPO compatibility on InternVL2.5.
}
\label{tab:hico_tpo}
\end{table*}

\paragraph{The compatibility between HiCo and TPO.} 
We further show HiCo and TPO are compatible and orthogonal about improving MLLMs in Table \ref{tab:hico_tpo}. Combing the training of HiCo and TPO together, it not only simplifies the overall learning with three stages, but also notably rises short and long video benchmark results as well as enabling NIAH of MLLMs. Compared with InternVL2.5-HiCo, {\modelname} gets non-trivial increases by 1.7, 3.5, 1.0, 1.0, and 1.3 points on MVBench, Percpetion Test, EgoSchema, LongVideoBench, and MLVU, respectively. The effectivenss on VideoMME is barely observed as the increase is only 0.2. It shows that TPO can improve both short and long video perception via classical vision supervision, but this benefits would vanish when the input videos are quite long (around and more than 15 minutes).

\section{Concluding Remarks}
This paper has introduced {\modelname}, a new version of video MLLM to improve the original perception and understaning ability through long and rich context (LRC) modeling. By focusing on improving context resolution—both in terms of length (memory) and fineness (focus)—{\modelname} enables current MLLMs to understand longer videos with detail emphasis. Our approach leverages direct preference optimization to transfer dense visual annotations to MLLMs, and employs adaptive hierarchical token compression for efficient spatiotemporal representation. Experimental results demonstrate that {\modelname} achieves the state-of-the-art performance on various video understanding benchmarks in around 7B model size, significantly increasing input video sequences by six times in length compared to the applied MLLMs. Furthermore, {\modelname} exhibits enhanced visual capabilities, including object tracking, showcasing the effectiveness of LRC in improving both fundamental vision tasks and higher-level reasoning. This work highlights the critical role of multimodal context resolution and richness in advancing MLLM capabilities and provides a promising direction for future research on improving MLLM performance through enhanced context processing.

\paragraph{Limitations.} 
While we demonstrate improved performance on long video sequences, the computational cost of processing such extended contexts remains significant. Further research is needed to explore more efficient learning techniques to reduce this overhead. Additionally, our current implementation primarily focuses on visual context properties. Extending LRC to reasoning-related areas, presents a promising avenue for future work and could further validate MLLM abilities.

\bibliographystyle{plainnat}
\bibliography{references} 

\end{document}